\documentclass[sigconf]{acmart}

\usepackage{booktabs,caption}
\usepackage[flushleft]{threeparttable}
\usepackage{hyperref}
\usepackage{color, colortbl}
\definecolor{LightCyan}{rgb}{0.88,1,1}
\usepackage{bigstrut}
\usepackage{subcaption}
\usepackage{natbib}
\setcitestyle{numbers,nosort&compress}
\usepackage{enumitem}
\usepackage{makecell}
\usepackage{soul}

\AtBeginDocument{%
  \providecommand\BibTeX{{%
    \normalfont B\kern-0.5em{\scshape i\kern-0.25em b}\kern-0.8em\TeX}}}

\setcopyright{acmcopyright}
\copyrightyear{2020}
\acmYear{2020}
\setcopyright{acmcopyright}
\acmConference[ACM GLSVLSI '20] {ACM Great Lakes Symposium on VLSI 2020}{September 7--9, 2020}{Virtual Event, China}
\acmBooktitle{ACM Great Lakes Symposium on VLSI 2020 (GLSVLSI '20), September 7--9, 2020, Virtual Event, China}
\acmPrice{15.00}
\acmDOI{10.1145/3386263.3406935}
\acmISBN{978-1-4503-7944-1/20/09}

\begin{document}

\title{Towards Systems Education for Artificial Intelligence: A Course Practice in Intelligent Computing Architectures}
\titlenote{A Microelectronic Systems Education Paper}

\author{Jianlei Yang}
\authornote{Corresponding author is Jianlei Yang. Email: jianlei@buaa.edu.cn}
\affiliation{
  \institution{SCSE, BDBC, Beihang University}
  \city{Beijing}
  \country{P. R. China}}
\email{jianlei@buaa.edu.cn}

\author{Xiaopeng Gao}
\affiliation{
  \institution{SCSE, Beihang University}
  \city{Beijing}
  \country{P. R. China}}
\email{gxp@buaa.edu.cn}

\author{Weisheng Zhao}
\affiliation{
  \institution{SME, BDBC, Beihang University}
  \city{Beijing}
  \country{P. R. China}}
\email{weisheng.zhao@buaa.edu.cn}

\thanks{This work is supported in part by the National Natural Science Foundation of China (61602022), State Key Laboratory of Software Development Environment (SKLSDE-2018ZX-07), CCF-Tencent IAGR20180101 and the 111 Talent Program B16001.}

\begin{abstract}

With the rapid development of artificial intelligence (AI) community, education in AI is receiving more and more attentions.
There have been many AI related courses in the respects of algorithms and applications, while not many courses in system level are seriously taken into considerations.
In order to bridge the gap between AI and computing systems, we are trying to explore how to conduct AI education from the perspective of computing systems.
In this paper, a course practice in intelligent computing architectures are provided to demonstrate the system education in AI era.
The motivation for this course practice is first introduced as well as the learning orientations.
The main goal of this course aims to teach students for designing AI accelerators on FPGA platforms.
The elaborated course contents include lecture notes and related technical materials.
Especially several practical labs and projects are detailed illustrated.
Finally, some teaching experiences and effects are discussed as well as some potential improvements in the future.

\end{abstract}



\maketitle

\section{Introduction}

In recent years, artificial intelligence (AI) has made great progress due to the various deep learning (DL) algorithms and applications.
Actually in scientific research community, AI's research has been extended from algorithm/application level to system level.
More and more attentions have been taken into system design problems in AI computing.
From the point of view of ACM Turning Lectures \cite{hennessy2019new,hinton19,lecun19}, the computer architecture's evolution is driven by AI's revolution. What's more, many interdisciplinary research topics will be inspired by AI and computer architectures/systems, such as the recent rising Conference on Machine Learning and Systems (MLSys) \cite{mlsys}.

Although AI technologies have been well developed in the scope of scientific research and industrial applications, more attentions should be taken into education scope in order to satisfy the urgent requirements of industrial technologies developments.
Nowadays the curriculum system in most of colleges or universities is relative old and lacks of enough novelty compared with the rapid development in technical stacks and social applications.
If most of the students are trained with lacking of enough system capabilities, they cannot meet the full-stack talents requirement of the AI industry.
Hence, the personnel training for AI community cannot lack of computing system hierarchies. Meanwhile, the education in computer system has to consider the AI-inspired architectures and systems.
In summary, the gap between education and industry has to be bridged by incorporating the computing architecture/systems and AI algorithms/applications.

\begin{table}
\caption{Related courses in ML architecture and system.}\label{tab:courses}
\begin{threeparttable}[t]
\begin{tabular}{|l|p{170pt}|}
    \hline
    \rowcolor{LightCyan}
    Affiliation & Course Name   \bigstrut \\ \hline
    UIUC  & Machine Learning in Silicon \footnotemark[1]    \bigstrut \\ \hline
    Beihang Univ.  & Intelligent Computing Architectures  	\bigstrut \\ \hline
    MIT  & Hardware Architecture for Deep Learning \footnotemark[2]  \bigstrut	 \\ \hline
    Stanford  & Hardware Accelerators for Machine Learning \footnotemark[3]  \bigstrut \\ \hline
    Univ. of CAS  & 	Intelligent Computing Systems \footnotemark[4]   \bigstrut \\ \hline
    UC Berkeley  & AI-Sys \footnotemark[5]   \bigstrut \\ \hline
    U. Washington & System for ML \footnotemark[6]  \bigstrut \\ \hline
    Georgia Tech. & Hardware Acceleration for Machine Learning \footnotemark[7]   \bigstrut \\ \hline
\end{tabular}
\begin{tablenotes}
\footnotesize
    \item[1] \url{https://courses.grainger.illinois.edu/ece598ns/fa2017/}
    \item[2] \url{https://www.eecs.mit.edu/academics-admissions/academic-information/subject-updates-spring-2019/6s0826888}
    \item[3] \url{https://cs217.stanford.edu/}
    \item[4] \url{http://novel.ict.ac.cn/aics/}
    \item[5] \url{https://ucbrise.github.io/cs294-ai-sys-sp19/}
    \item[6] \url{https://dlsys.cs.washington.edu/}
    \item[7] \url{http://tusharkrishna.ece.gatech.edu/teaching/hml\_s19/}
\end{tablenotes}
\end{threeparttable}
\end{table}

There have been many courses in AI algorithms and applications as well as the traditional computer architecture and systems.
And the related resourceful course experiments have a very good training effect.
However, it lacks of sufficient and timely exploration in the teaching AI-inspired systems and architectures.
If AI education only focuses on the higher level topics, the students will lack of the vision and capability to understand the system performance issues on AI computing.
Therefore, with the goal of cultivating talents in the full stack of AI and starting from cultivating students' computer system capabilities, we will introduce the demand of computing system in the AI era into the curriculum education. The main goals of this course exploration are listed as following:
\begin{itemize}[leftmargin=8pt,topsep=4pt,itemsep=0pt,parsep=0pt]
    \item \textit{Bridge the education gap between AI and computing architecture/system:} broaden the students' research vision and towards full-stack development skills in AI-related techniques.
    \item \textit{Introduce emerging research topics on AI-inspired architecture/system into college courses:} motivate students to learn and study computer architecture/system instead of only focusing on the algorithms/applications in AI era.
\end{itemize}

\section{Background and Related Courses}

The Turing Lecture \cite{hennessy2019new} points out that domain-specified computing architecture is an important trend in the future customization computing.
Especially in recent years, how to design high-performance and energy-efficient domain-specified architectures have become as very popular research directions due to the rapid growth of AI applications on the urgent requirements for computing \cite{cong2010customizable}.

\begin{figure}[t]
    \centering
    \includegraphics[width=0.42\textwidth]{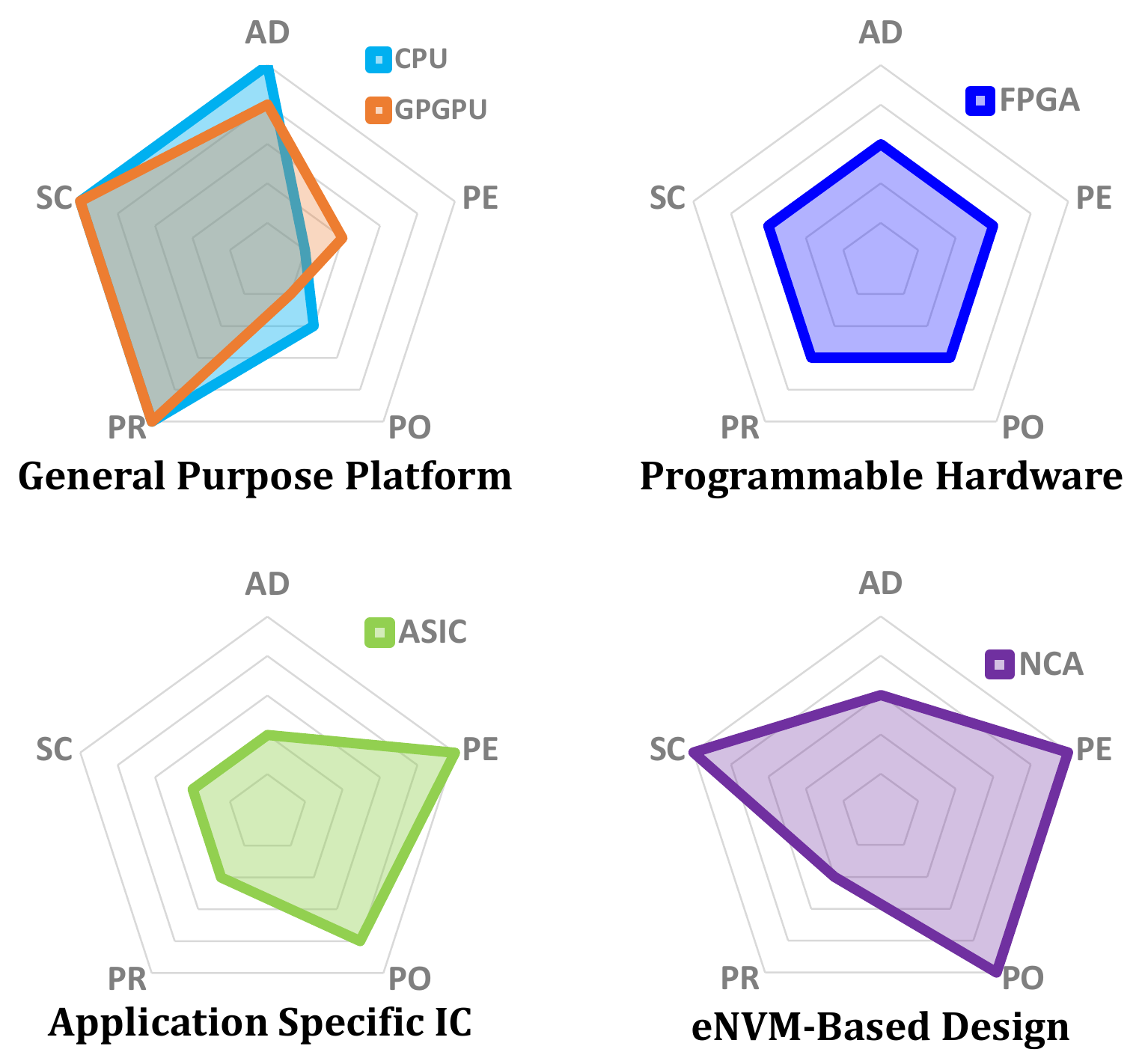}
    \caption{Comparison among different design platforms \cite{mahajan2016tabla,park2017scale}, where AD means Adaptivity, PE means Performance, PO means Power Efficiency, PR means Programmability, SC means Scalability.}
    \label{fig:platform}
\end{figure}

As shown in Figure \ref{fig:platform}, different kind of computing platforms are compared in terms of adaptivity (AD), performance (PE), power efficiency (PO), programmability (PR) and scalability (SC) \cite{mahajan2016tabla,park2017scale}.
General computing platforms, such as CPU or GPU, have good advantages in adaptivity, scalability and programmability, but their performance and power efficiency are greatly limited for specialized computing purpose.
Application specified platforms, such as NPUs, could achieve very high performance, and power efficiency, but are very poor in adaptivity and scalability.
Reconfigurable copmuting platforms, such as FPGAs, have a good balance in performance, efficiency and adaptivity, but do not have obvious advantages in each dimension while their advantages are convenient for rapid design iteration.
In addition, there are some emerging NVM-based neuromorphic computing platforms, which have good advantages in scalability, performance and efficiency, but also have obvious limitations in adaptivity and programmability due to lack of available development tools or methodologies. 
As a result, it is difficult to find a computing platform that is excellent in all dimensions for domain customized architecture and chip design.
In this course, the idea of domain-specified architecture design is introduced and some developing methodologies are provided to train students for design skills improvements. 
In terms of platform consideration, FPGA is adopted in this course for rapid development iteration.

Recently, some relevant courses have appeared in many universities as shown in Table \ref{tab:courses}. Most of these courses are focusing on AI-inspired architecture and system education, which have received great attention and good effects.
Prof. Naresh R. Shanbhag in UIUC offers a course named \textit{Machine Learning in Silicon} since Fall 2017. This course aims to teach students how to design machine learning algorithms in chips directly, which mainly focuses on circuit and system implementations \cite{kang2019energy}.
Prof. Vivienne Sze and Joel Emer in MIT offer a course named \textit{Hardware Architecture for Deep Learning} since Fall 2017. This course is more likely a tutorial to present their researches in DL architectures \cite{sze2017efficient}.
Prof. Ardavan Pedram and Kunle Olukotun in Stanford Univ. offer a course named \textit{Hardware Accelerators for Machine Learning} since Fall 2018. This course introduces many latest hardware accelerators for machine learning applications.
Dr. Yunji Chen in Univ. of CAS offer a course named \textit{Intelligent Computing Systems} since Fall 2018. This course mainly promote their research achievements of DianNao architectures as well as their Cambricon chips and ecosystems \cite{cambrican}. They have published a relevant teaching textbook and provided resourceful experimental materials. This course has been promoted to several universities in China, which has a great influence in education community.
Since 2019, there have been several courses focused on ML system level.
Prof. Ion Stoica and Joseph E. Gonzalez in UC Berkeley offered a course named \textit{AI-Sys}. It describes the latest trends in systems designs to better support the next generation of AI applications, and applications of AI to optimize the architecture and the performance of systems.
Prof. Tianqi Chen in Univ. of Washington offered a course named \textit{System for ML} since Spring 2019. This course is designed to fill the gap in how to build and optimize these deep learning specified systems.
Prof. Tushar Krishna in Georgia Tech. offered a course named \textit{Hardware Acceleration for Machine Learning} since Spring 2019. This course present various development resources that can enable researchers and practitioners to quickly get started on DNN accelerator design, and highlight important benchmarking metrics and design considerations that should be used for evaluating the rapidly growing number of DNN hardware platforms being proposed in academia and industry.
In summary, these courses listed above provided a very useful early exploration by introducing circuit/chip/architecture/system levels considerations on how to build and optimize deep learning computing systems.

Our course, named \textit{Intelligent Computing Architectures}, mainly focused on the architectures explorations for deep learning tasks since Fall 2017.
Most attentions are paid to present recent advances in architectures towards the goal of enabling efficient processing of DNNs.
This course requires the students to have a wide scope of knowledge and strong practical capabilities which is very challenging to finish the experiments and projects.
But the students' system developing capabilities will be greatly improved once they have gone through the intensive training in this course.

\section{Course Design and Implementation}

\subsection{Course Prerequisites}

\begin{figure}[t]
    \centering
    \includegraphics[width=0.49\textwidth]{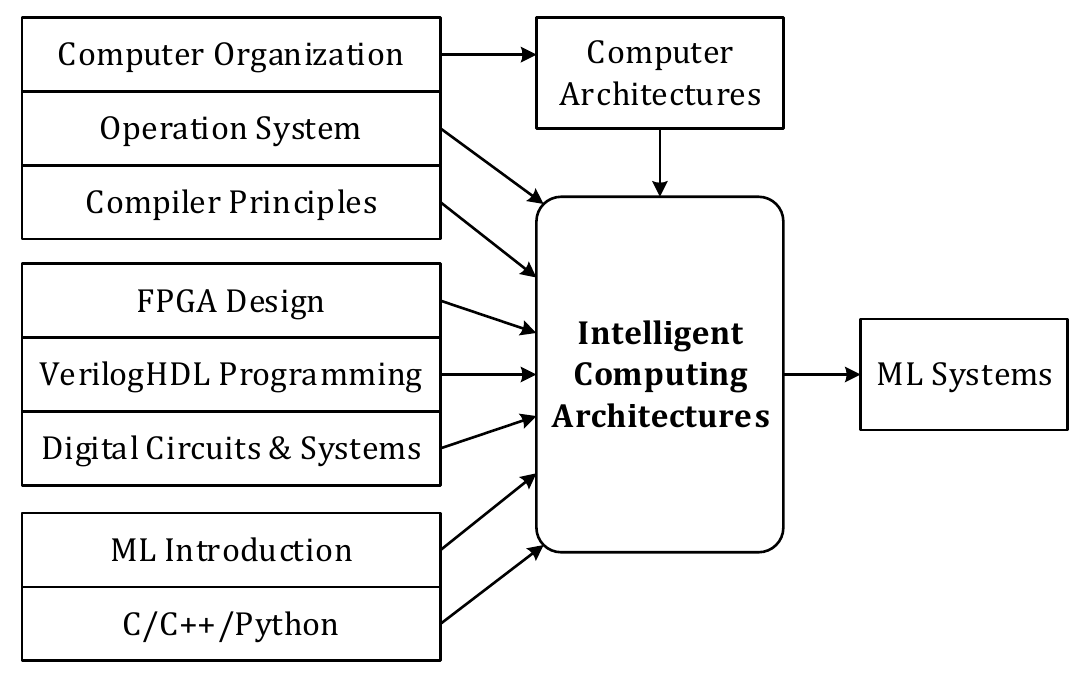}
    \caption{Course prerequisites and following courses.}
    \label{fig:prerequisites}
\end{figure}

This course presents how to design domain-specific hardware architectures, which requires students to learn and practice both hardware and software knowledge. The prerequisites include three parts: computer architecture and system, hardware design, and algorithm related background. For the system part, students are fist required to learn computer architectures to understand how to build instructions set and optimize hardware implementations. Also some background of computer compiler are required for mapping ML algorithms as instructions. For the hardware design part, students are required to learn digital circuit \& system, Verilog HDL language, as well as FPGA development techniques for hardware implementations. For the algorithms related background, students are required to know some ML introductions and know well about programming techniques, including C, C++ and Python programming. With these backgrounds, students in this course could learn how to design ML architectures and implement them into FPGA platform with the support of dataflow instructions compilation. Based on these achievements, students could try to learn how to design and optimize ML systems with the architectural level considerations.

\subsection{Teaching Objectives}

In this course, some emerging research are introduced into classroom so that it involves many basic knowledge and skills, especially the software-hardware co-design methodologies. This course requires students to have very strong learning capabilities and operational capabilities.
Based on several typical applications in the field of intelligent computing, this course introduces the design and implementation methodologies of mapping software/algorithms to hardware architectures. From this course, students could understand the working principles of intelligent computing in the point of view of computer architectures. With the course experimental training, students could obtain the skills of designing and implementing domain-specified hardware for intelligent algorithms. This course requires students to learn and know how to independently design and prototype several typical intelligent computing accelerators. Their engineering opinions of heterogeneous system design could be built for extending further research vision. The main teaching objectives of this course could be listed as following:
\begin{enumerate}[leftmargin=10pt,topsep=5pt]
    \item Know some typical intelligent computing algorithms, including machine learning, data mining, etc. Know some basic problems and techniques in hardware acceleration area.
    \item Master some design and optimization methodologies of digital systems. Know how to map intelligent algorithms into hardware architectures.
    \item Familiar with development tools, such as Verilog, Modelsim, Vivado, etc. Familiar with deep learning frameworks, such as Caffe, PyTorch, TensorFlow, etc.
    \item Know how to access and utilize technical documentations.
    \item Master embedded system design and optimization techniques.
    \item Obtain modelling, design and analysis skills of intelligent computing systems, as well as the innovation capabilities in solving practical problems through acquired knowledge and skills.
\end{enumerate}

\subsection{Teaching Contents and Practices}

\begin{table}[b]
\caption{Teaching contents and class hours in this course, while total teaching hours are 32 hours in 16 weeks (2 hours for each week, total 16 weeks).}\label{tab:coursecontents}
\begin{center}
\begin{tabular}{|p{180pt}|p{38pt}|}
    \hline
    \rowcolor{LightCyan}
    Teaching Contents & Schedule \bigstrut  \\ \hline
    \textbf{Introduction to intelligent computing architectures:} the motivations to learn this course, the teaching scope and required learning capabilities.  &  \makecell{2 Hours \\ Week 1} \bigstrut \\ \hline
    \textbf{Mainstream computing models and methods:} DNNs and graph computing methods, especially targeted for hardware design issues. &  \makecell{4 Hours \\ Week 2-3} \bigstrut \\ \hline
    \textbf{Domain-specified architecture design methodologies:} design principles and development flow from algorithms to hardware, targeted on FPGA or ASIC implementations. &  \makecell{4 Hours \\ Week 4-5}  \bigstrut \\ \hline
    \textbf{Compiling or mapping methodologies:} system modeling, functionalities partitioning, dataflow mapping and scheduling, performance and efficiency optimization techniques. &  \makecell{6 Hours \\ Week 6-8} \bigstrut \\ \hline
    \textbf{DNN accelerators design:} algorithms evaluations, dataflow architectures, hardware design with Verilog/HLS implementations targeted on Xilinx Zynq FPGA platforms. & \makecell{10 Hours \\ Week 9-13}  \bigstrut \\ \hline
    \textbf{ASIC implementation flow:} a brief introduction including behavior description, logic synthesis, physical implantation, system verification, timing analysis and optimization, etc. &  \makecell{4 Hours \\ Week 14-15} \bigstrut  \\ \hline
    \textbf{Application perspectives of hardware accelerators:} an outlook of practical applications in big data analytic, DL, CV, robotics, etc. &  \makecell{2 Hours \\ Week 16}  \bigstrut \\ \hline
\end{tabular}
\end{center}
\end{table}

\begin{table}[b]
\caption{Labs and projects design in this course.}\label{tab:labsandprojects}
\begin{center}
\begin{tabular}{|c|p{7cm}|}
    \hline
    \rowcolor{LightCyan}
    No. & Lab or Project Details (\ul{\textit{Week Schedule}}) \bigstrut \\ \hline
    Lab 1 & \textbf{Deep neural networks:} learn and write C++/Python code for some deep neural networks to understand the learning principles and network structures. \ul{\textit{Week 3}} \bigstrut \\ \hline
    Lab 2 & \textbf{Deep learning frameworks:} run deep neural networks with Caffe/TensorFlow/PyTorch frameworks, evaluate the model accuracy and performance. \ul{\textit{Week 4}} \bigstrut \\ \hline
    Lab 3 & \textbf{Zynq FPGA development:} learn Zynq FPGA development flows, including Vivado design suite, IP usage, AXI bus protocol, PS and PL co-design, etc. \ul{\textit{Week 5-6}} \bigstrut \\ \hline
    Lab 4 & \textbf{MAC module design with Verilog:} implement Verilog modules for matrix multiplication and accumulation, validate on PL part of Zynq FPGA. \ul{\textit{Week 7-8}} \bigstrut \\ \hline
    Prj 1 & \textbf{MAC design on Zynq FPGA:} implement MAC modules on PL part and perform validation with the controlling of PS part in Zynq FPGA. \ul{\textit{Week 9-10}} \bigstrut \\ \hline
    Prj 2 & \textbf{LeNet design on Zynq FPGA:} implement the data path and controller to run LeNet on Zynq FPGA, only on-chip BRAMs are exploited, where the PS part is in charge of data input/output and dataflow controlling, the PL part is in charge of computations. \ul{\textit{Week 11-13}} \bigstrut \\ \hline
    Prj 3 & \textbf{VGGNet design on Zynq FPGA:} VGGNet implementation which is similar to LeNet, where the off-chip DRAM is also utilized for buffering the intermediate data as well as the on-chip BRAMs. \ul{\textit{Week 14-16}} \bigstrut \\ \hline
\end{tabular}
\end{center}
\end{table}

As listed in Table \ref{tab:coursecontents}, this course consists of 32 teaching hours and mainly includes three parts: basic preparation, accelerator design, further explorations.
In the basic preparation part (6 hours), intelligent computing architectures are first introduced as well as the mainstream computing models and methods. It motivates students' interests to learn this course and know some research and technical backgrounds in related communities.
Meanwhile, some DNNs are described in the class and students evaluate some DNN models to understand their computational dataflow.

In the accelerator design part (20 hours), system architectures and dataflow mappings are detailed presented.
The basic FPGA design flows are introduced for development in Zynq platforms. Some resourceful Zynq materials, including documentations, design examples, application notes, etc., are provided to students for learning and developing.
Then, some popular architectures of DNN accelerators are discussed, which are arisen from latest research papers or chips in recent 6 years, such as DianNao\cite{chen2014diannao}, Eeyriss\cite{chen2016eyeriss}, TPU\cite{jouppi2017datacenter}, etc.
Furthermore, a simple DNN architecture is provided as a reference to our labs and projects. Several typical operations are defined as instructions or pseudo-instructions to support our architectures for running whole DNN networks.
Meanwhile, dataflow compiling or mapping techniques are introduced. With the predefined ISA, DNN dataflow is translated as instructions which could run in the designed architectures. Some optimization techniques could be involved in this process to improve the accelerator's performance or energy efficiency.

In the further explorations part (6 hours), ASIC implementation flow is introduced and accelerators' applications are discussed.
Beyond the implementations on FPGA platforms, some ASIC design flows could broaden the students' research or technical visions.
Also some further perspectives of DNN accelerators' applications could motivate students to learn and explore potential possibilities in this emerging research direction.

Besides the teaching contents in the class, several labs and projects are provided as shown in Table \ref{tab:labsandprojects}. Lab 1-4 help students to understand the DNN algorithms and frameworks as well as the FPGA developing skills. Based on these four labs, students are required to build a DNN accelerator step by step among project 1 to project 3. Aiming to finish these tasks, students are usually required to spend much more times than in class. Lab 1 is provided to make students familiar with the frequently used DNN algorithms. Lab 2 aims to evaluate DNNs with deep learning frameworks, which is important to balance the model accuracy and performance/efficiency in architectural explorations. Lab 3 provides a chance to learn Zynq FPGA developing skills while many design examples are provided as a reference. Lab 4 requires students to design a multiply-and-accumulation (MAC) module with Verilog, which should be validated in PL part of Zynq FPGA. It should be not difficult for students who have learnt Verilog and digital systems design courses before. These four labs aims to make students familiar with DNN fundamentals and hardware design platforms.

These labs and projects are provided progressively towards a practical DNN accelerator design. The main target of this course aims to design DNNs accelerator in FPGA which is covered by the remaining three projects. In project 1, the designed MAC module in Lab 4 is integrated with ARM processor as main controller in PS part. Vivado SDK is utilized for software developing while the PS-PL communication mechanisms should be well exploited for software-hardware co-design. Project 2 aims to design a LeNet on Zynq FPGA while all computing and data buffering tasks could be finished within on-chip resources. Since the LeNet model size is relatively small, such as less than 1Mb, the model weights and intermediate data could be handled just with on-chip block RAMs. Project 3 is our final project, which targets to design a VGGNet on Zynq FPGA. The VGGNet model size is very large so that we have to store most of weights and intermediate data in off-chip DDR memory.

\begin{figure}[t]
    \centering
    \includegraphics[width=0.49\textwidth]{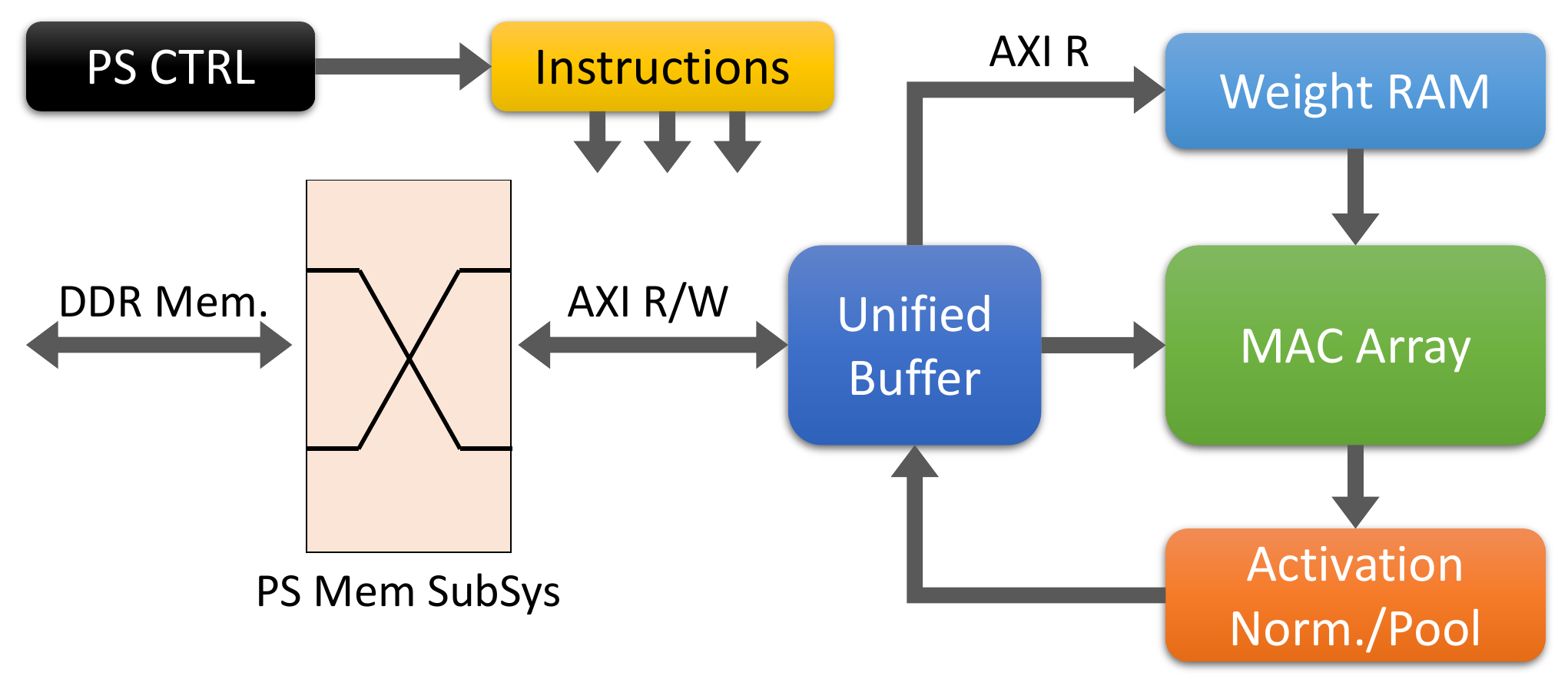}
    \caption{A typical architecture reference for DNN accelerator design, where three factors are mainly considered: data transfer bandwidth, MAC array scale, on-chip buffer size.}
    \label{fig:architecture}
\end{figure}

\begin{figure}[t]
    \centering
    \includegraphics[width=0.49\textwidth]{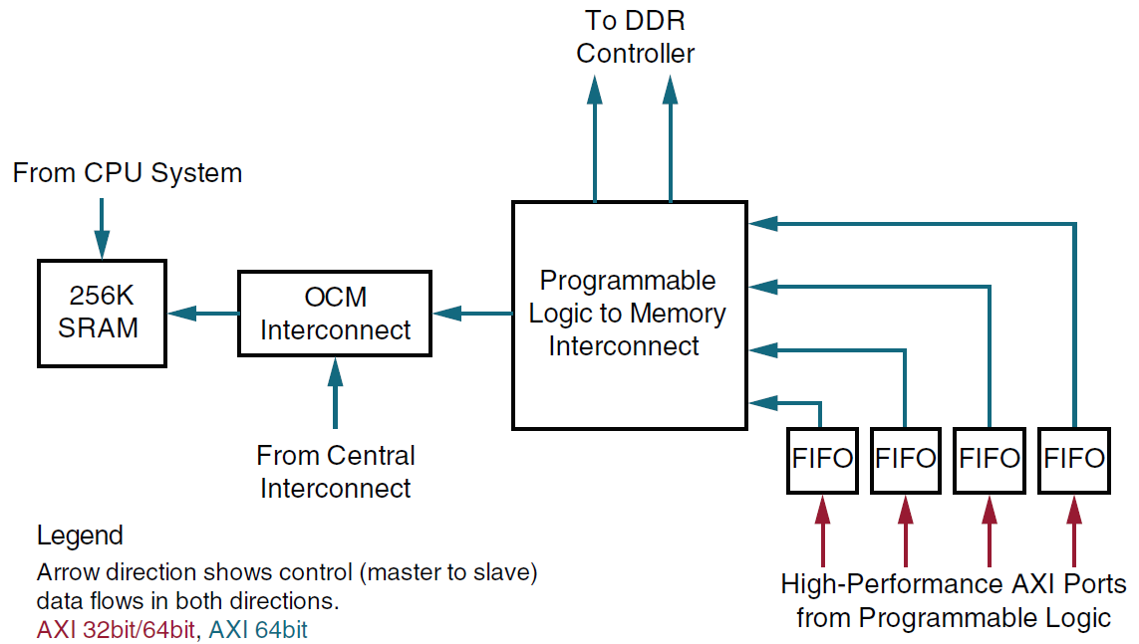}
    \caption{Data access from PL part to DDR via the PL Interface to PS Memory Subsystem \cite{ds190}.}
    \label{fig:pl2ddr}
\end{figure}

A typical architecture is provided as shown in Figure \ref{fig:architecture}, which is a simple reference for students to design DNN accelerators. The original data including network models and images are pre-stored in on-board SD card. The PS part in Zynq is in charge of system controlling, including data input/output, data caching and computing, etc. The memory system includes CPU cache, OCM, block RAMs, off-chip DRAM, etc. Data access between PL and PS is maintained by a memory interconnect block as shown in Figure \ref{fig:pl2ddr}. Especially when DRAM is utilized for data accessing by PL part, it requires to use the PL interface to PS memory sub-system. Once system booting is finished, DNN models are loaded into DRAM and partially stored in unified buffer (in BRAMs) while weights are stored in weight RAMs. MAC array is the core computation part to finish multiplication and accumulation. After computation of MAC array, a post-processing part is utilized to compute activation, normalization, pooling, etc.

DNN computation dataflow is compiled as instructions to control typical operations, including computing, data caching, data transfer, etc. Several typical instructions includes:
\begin{itemize}[leftmargin=10pt,topsep=5pt]
    \item \texttt{Read\_Host\_Memory}: read data from PS to unified buffer.
    \item \texttt{Read\_Wights}: read weights from DRAM to weight FIFO.
    \item \texttt{MatrixMultiply/CONV}: convolutional computation.
    \item \texttt{Activate}: activation or pooling computation.
    \item \texttt{Write\_Host\_Memory}: write results from unified buffer to DRAM.
\end{itemize}

\begin{figure}[t]
    \centering
    \includegraphics[width=0.49\textwidth]{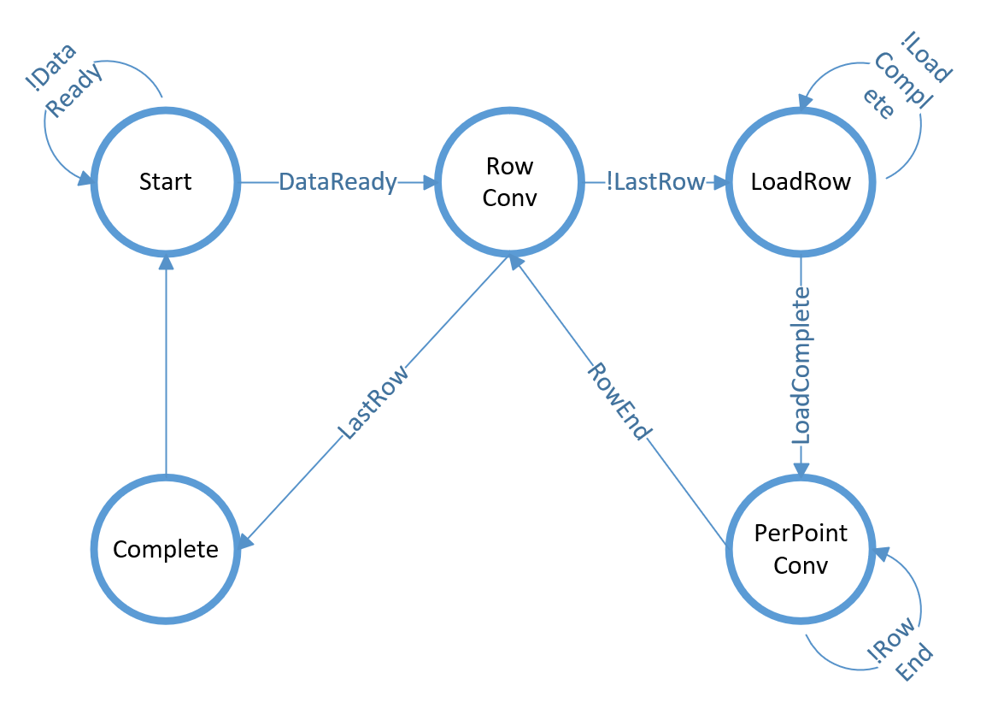}
    \caption{Finite state machine description for a typical computing dataflow.}
    \label{fig:fsm}
\end{figure}

\begin{table}[t]
\caption{Resources utilization comparison between Google TPU and our referred Na\"iveTPU design.}\label{tab:naivetpu}
\begin{center}
\begin{tabular}{|c|c|c|}
    \hline
    \rowcolor{LightCyan}
    Resources & Google TPU & Our Na\"iveTPU \bigstrut  \\ \hline
    Matrix Multiply Unit & 256 $\times$ 256 & 32 $\times$ 32  \bigstrut \\ \hline
    Accumulators RAM &  4K $\times$ 256 $\times$ 32b   &  4K $\times$ 32 $\times$ 32b  \bigstrut \\ \hline
    Unified Buffer &  96K $\times$ 256 $\times$ 8b  &  16K $\times$ 32 $\times$ 8b   \bigstrut \\ \hline
\end{tabular}
\end{center}
\end{table}

The DNN computation dataflow is processed by a row-by-row manner while the finite state machine (FSM) is described as Figure \ref{fig:fsm}. It starts from data preparation, i.e., data have been pre-loaded into unified buffer or weight RAM. If the required data have been ready for each row convolutions, they are sent to MAC array for \texttt{CONV} computing. In our course, a typical systolic array is provided for \texttt{CONV} computing. As shown in Table \ref{tab:naivetpu}, our utilized Na\"iveTPU is a reduced version of Google TPU. The MAC array scale is reduced according to the available logic resources in our utilized Zynq chip. Unified buffer, FIFOs, and RFs are also determined by the available storage resources in our utilized Zynq chip.

\begin{figure}[t]
\subfloat[AX7020 board]{\includegraphics[width=0.22\textwidth]{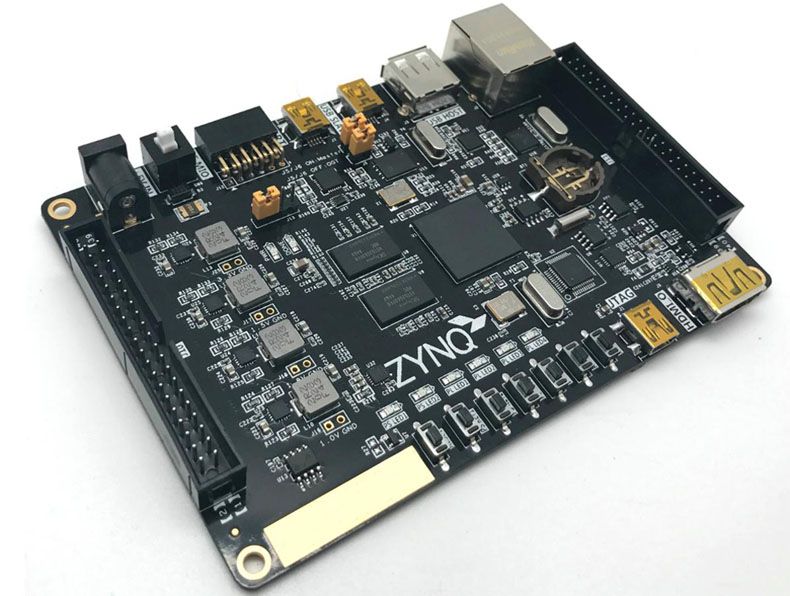}} 
\subfloat[AX7020 diagram]{\includegraphics[width=0.24\textwidth]{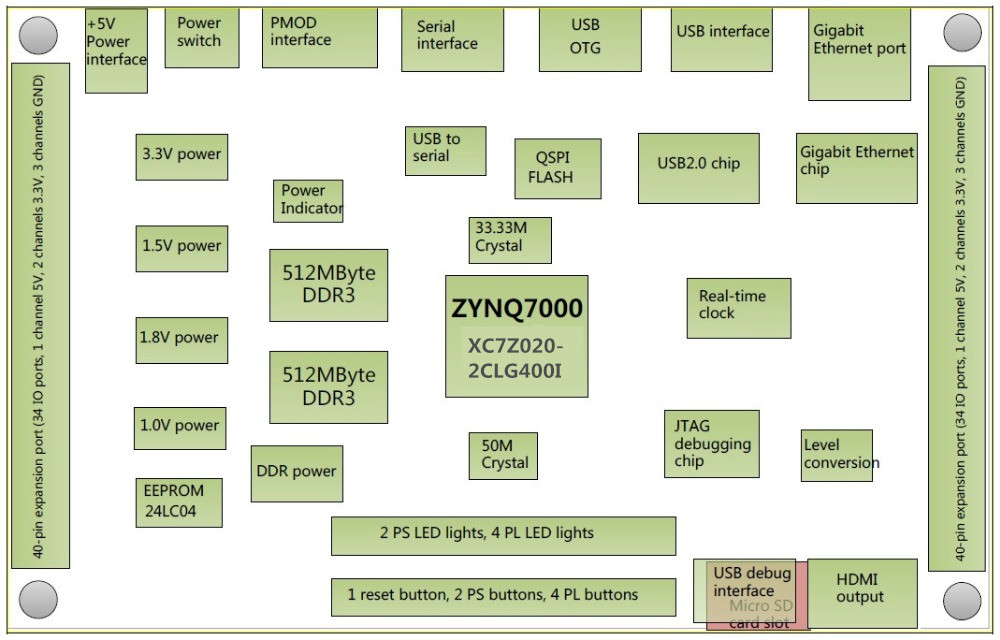}}
\caption{The utilized AX7020 Zynq FPGA board \cite{ax7020} in this course.}
\label{fig:ax7020}
\end{figure}

For course labs and projects, we provide a Zynq FPGA board for each student as shown in Figure \ref{fig:ax7020}. The utilized board is named AX7020, with a Zynq chip of XC7Z0202CLG400I, including a dual Cortex-A9 ARM core (PS), 4.9Mb BRAM, and on-board 8GBit DDR3 SDRAM. For running LeNet, on-chip BRAM is enough for data caching. But for VGGNet, off-chip DRAM is required to store most of the model weights and intermediate data. Students are required to understand how to access DDR from PL via the interface to PS memory sub-system. Some important demos or reference designs are provided to help students to finished their project, including: SD card access API, DDR memory access API, AXI controller, BRAM usage, UART uasge, etc. All of these examples are included in the AX7020 suite with comprehensive code examples and documentations.

\section{Teaching Practice and Discussion}

In this section, we mainly discuss some key observations we have acquired from teaching practice in each Fall semester since 2017 to illustrate the efficiency and feasibility of our course in motivating students to study AI-inspired architectures.

This course requires students to spend many time in reading documentations and system developing. These efforts are necessary to well understand the architectures and FPGA developing flows.
From our observations, most of the students are very interested in learning these knowledge and willing to spend many efforts to finish our labs/projects.
All students could finish the lab 1-4 to project 1 well since these tasks are fundamental requirements. About 80\% students could finish the LeNet design project and demo their functions to us.
The remaining 20\% students usually have some difficulties in timing optimization and software-hardware co-design stage.
For VGGNet design, about 60\% students could demo their full functionalities. Some of them even could achieve a relatively high performance when evaluated on ImageNet dataset.
The remaining 40\% students usually could not handle the complicated FSMs since there are many data transfer and data reuse strategies for VGGNet computations.

From our observations, the most excellent students (about 10\%) in this course could design a very robust DNN accelerator in Zynq FPGA while their performance and efficiency could be very close to the reported results in some early research papers \cite{qiu2016going}.
Meanwhile, a survey from the students in this course of first two years shows that the top excellent students (about 10\%) choose deep learning accelerators as their research directions in the postgraduate study.
It indicates that most of the students in this course are very interested in our teaching even though it is very challenging and has many difficulties.
These students have learnt lots of knowledge in DNN architectures and achieved strong system developing skills in this course.
What's more, their inspired interests and confidence enable them contribute to this research field in the future.

\section{Limitations and Future Work}

\textbf{Course Workloads.} From our course practice in previous three years, we have noticed that students have a very heavy workloads both in leaning new knowledge and mastering new developing skills.
We plan to suggest students to further improve their FPGA developing skills and practice more DNN algorithms in their pre-acquired courses. 
Hence, we could introduce more architectural details in DNN accelerators. And furthermore, students could try to design more complicated architectures and optimize their performance and efficiency.
These achievements could provide a stronger background for their future research.

\noindent
\textbf{System Capabilities.} This course mostly focuses on architectural design issues while lacking of enough system level implementations.
Aiming to enable students to be qualified with full-stack techniques, we would like to further offer a new course from Fall 2020, named \textit{Intelligence Computing Systems}, to provide more system level techniques. This new course will mostly focus on compilers of mapping DNN algorithms into specified architectures, as well as the related optimization methodologies.

\section{Conclusions}

In this paper, we present a new course focused on intelligent computing architectures. Based on recent researches, this course introduces lots of domain-specified architectures and provides resourceful experimental materials. These training tasks are very challenging for students but they are motivated to study and practice deeply.
We believe that such practice could be helpful to cultivate students towards full-stack system developing capabilities in artificial intelligence era.

\bibliographystyle{unsrt}
\bibliography{glsvlsi}

\end{document}